\documentclass[letterpaper]{article} 
\usepackage[preprint]{aaai2027}
\usepackage{amsmath}
\usepackage{amssymb}
\usepackage[hyphens]{url}  
\usepackage{graphicx} 
\usepackage{natbib}  
\usepackage{caption} 
\usepackage{algorithm}
\usepackage{algorithmic}

\usepackage{newfloat}
\usepackage{listings}
\DeclareCaptionStyle{ruled}{labelfont=normalfont,labelsep=colon,strut=off} 
\floatstyle{ruled}
\newfloat{listing}{tb}{lst}{}
\floatname{listing}{Listing}

\usepackage{booktabs}

\title{Identity-Aware Human-Object Interaction Motion Captioning}
\author{
       Yiming Wang, Yonghao Dang, Huilai Li, Jiawei Tu, Jianqin Yin
}
\affiliations{
}

\begin{document}

\maketitle

\begin{abstract}
Existing human-object interaction (HOI) motion captioning methods typically describe \emph{what happens} while referring to the subject using generic terms such as ``a person'' or ``someone'', without grounding the caption in subject identity. To address this limitation, we introduce \textbf{Identity-Aware Human-Object Interaction Motion Captioning} task. This task requires each generated caption to specify both the subject identity and the corresponding HOI motion. For example, the model generates ``Sub\_ID lifts the chair'' rather than ``A person lifts the chair''. For this task, we design identity-aware HOI motion captions based on the BEHAVE and InterCap datasets. We further propose \textbf{ID-HOINet}, which learns from multi-view videos while supporting single-view identity-aware HOI motion caption generation. ID-HOINet contains two core components: Multi-View Identity-Motion Learning Module (\textbf{MVIML}) and Two-Stage Caption Rewriting Strategy (\textbf{TSCR}). {MVIML} learns from multi-view videos by modeling dependencies across temporal stages and camera viewpoints, capturing identity and interaction motion features. At inference, the {TSCR} first retrieves the subject identity and generates identity-agnostic HOI motion captions. TSCR then rewrites these captions with the predicted identity to produce the final identity-aware HOI motion captions. Experiments demonstrate that ID-HOINet achieves state-of-the-art performance. Code will be released upon acceptance.
\end{abstract}


\section{Introduction}

Human-Object interaction (HOI) motion captioning aims to describe the interaction relation between a human and an object using natural language. Recent captioning approaches have achieved significant progress in formulating coherent natural-language captions from visual motion inputs \cite{tang2021clip4caption,lin2022swinbert,yang2023vid2seq,wang2024omnivid}. However, these methods typically generate identity-agnostic HOI motion captions, such as ``A person picks up a box'' or ``Someone sits on a chair''. Such captions describe \emph{what interaction occurs} but do not specify \emph{who performs it}. This limitation is critical in identity-dependent scenarios, such as personalized rehabilitation monitoring \cite{li2024finerehab,kryeem2023personalized,hakim2019mal} and multi-subject motion analysis \cite{doering2022posetrack21,wang2020combining,liu2025motions}, where the observed motion must be associated with the correct individual.

Recently, multiple identity-aware video captioning techniques have been introduced to link captions with their respective subject identities within cinematic and sports videos
\cite{park2020identity,raajesh2024micap,xi2025player}. However, these methods mainly describe subjects performing actions that involve a fixed object or no explicit object interaction. In complex human-object interaction scenarios, the captions generated by these methods fail to explicitly associate the subject identity with the corresponding HOI motion.

\begin{figure}[!t]
\centering
\includegraphics[width=\columnwidth]{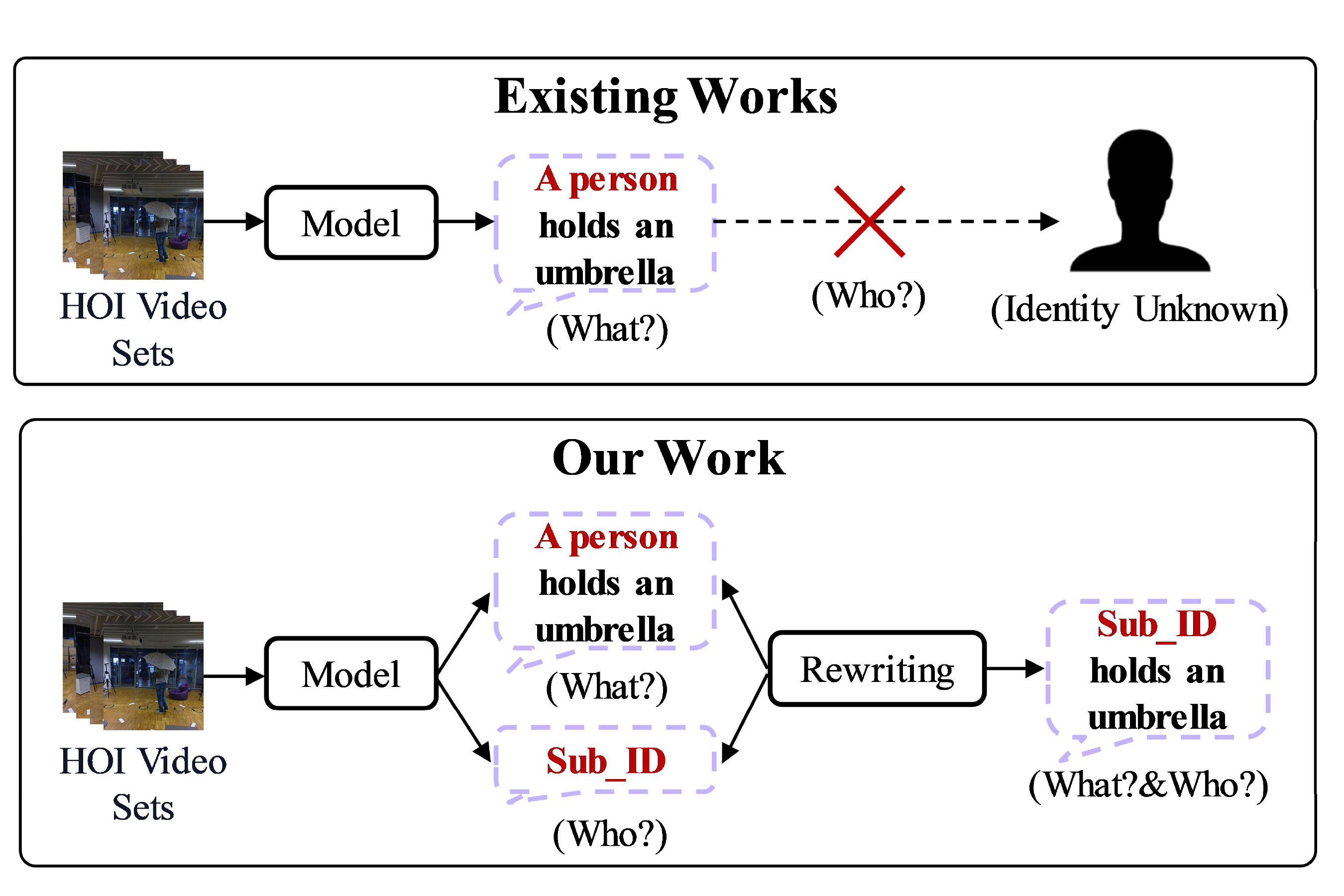}
\caption{
Motivation of identity-aware HOI motion captioning. Existing methods generate identity-agnostic captions that only describe \textit{what} happens, while our work further identifies \textit{who} performs the motion, enabling \textit{who-and-what} motion captioning.
}

\label{fig:motivation}
\end{figure}

Accordingly, as illustrated in Figure~\ref{fig:motivation}, we propose \textbf{Identity-Aware Human-Object Interaction Motion Captioning} task. To the best of our knowledge, this is the first study to extend identity-aware video captioning to the human-object interaction motion task. Unlike prior work focused on movies and sports videos, our generated caption explicitly associates the subject identity with the corresponding HOI motion. Given a multi-view HOI video, this task requires the model to generate an identity-aware HOI motion caption such as ``Sub\_ID lifts the chair'', rather than the identity-agnostic caption ``A person lifts the chair''. 
To construct an identity-aware HOI motion captioning dataset, we reorganize the multi-view video data from BEHAVE \cite{bhatnagar2022behave} and InterCap \cite{huang2024intercap}. We divide the synchronized videos into temporally complete HOI segments and provide each segment with an identity-aware HOI motion caption annotation that explicitly associates the observed subject identity with the corresponding interaction motion.
The annotated HOI segments are then partitioned by subject identity into disjoint training and test sets, ensuring that no test subject appears in the training identity set. During testing, the model is required to determine the identity of each subject by matching a single-view query video to reference subject samples in a gallery \cite{zheng2016mars} and incorporates the retrieved identity into the generated HOI motion caption.

To this end, we propose \textbf{ID-HOINet} for identity-aware human-object interaction motion captioning. ID-HOINet employs a Multi-View Identity-Motion Learning Module (\textbf{MVIML}) to learn identity and interaction motion features from multi-view videos. During inference, ID-HOINet then adopts a Two-Stage Caption Rewriting Strategy (\textbf{TSCR}) that first determines the subject identity and generates an identity-agnostic HOI motion caption, and subsequently combines them through a pretrained Caption Rewriting Decoder to produce the final identity-aware HOI motion caption.

Extensive experiments demonstrate that ID-HOINet achieves state-of-the-art performance on the proposed task, showing its effectiveness in associating subject identities with the corresponding HOI motion captions.

The main contributions of this work are summarized as follows:
\begin{itemize}
    \item We introduce \textbf{Identity-Aware Human-Object Interaction Motion Captioning}, a new task that requires the generated caption to explicitly associate subject identity with the corresponding HOI motion.

\item We propose \textbf{ID-HOINet}, which comprises two core components: the Multi-View Identity-Motion Learning Module (\textbf{MVIML}) and the Two-Stage Caption Rewriting Strategy (\textbf{TSCR}). MVIML captures identity and motion cues from multi-view videos. At test time, {TSCR} first matches the query subject to a reference gallery and generates an identity-agnostic HOI motion caption, then integrates the retrieved identity via a pretrained Caption Rewriting Decoder to produce the final identity-aware HOI motion caption.

    \item Extensive experiments demonstrate that ID-HOINet achieves state-of-the-art performance under the proposed evaluation setting, validating the effectiveness of the proposed framework.
\end{itemize}

\section{Related Work}

\subsection{Human-Object Interaction Motion Captioning}

Human-object interaction (HOI) motion captioning aims to generate a caption that expresses how a subject interacts with an object. To capture the visual information required for HOI motion features, related methods focus on modeling object interactions, human motion, and general video understanding. For interaction modeling, SINet-Caption jointly models HOI relations among multiple objects for fine-grained caption generation~\cite{ma2018attend}, while SAVCHOI incorporates HOI features into dense captioning for surveillance videos~\cite{mittal2022savchoi}. For motion modeling, existing methods introduce pose-based angular features~\cite{zhao2025exploring} or human-mesh motion features~\cite{song2025towards} to capture fine-grained body movements. General video captioners further improve video-to-language generation through CLIP-based visual-language knowledge~\cite{tang2021clip4caption}, sparse video attention~\cite{lin2022swinbert}, or unified sequence modeling~\cite{yang2023vid2seq,wang2024omnivid}.

These methods generally generate identity-agnostic HOI motion captions that specify \emph{what HOI motion is performed} but not \emph{who performs it}. In contrast, our task requires the generated caption to explicitly associate the subject identity with the corresponding HOI motion.

\subsection{Identity-Aware Video Captioning}

Identity-aware video captioning extends conventional video captioning by requiring the generated descriptions to not only express the observed events but also associate their corresponding subject identities. Existing studies mainly investigate this problem in films and sports events. Specifically, Park et al.~\cite{park2020identity} study identity-aware multi-sentence movie description using a two-stage framework that first generates anonymous descriptions and then resolves character identities across multiple video segments. MICap~\cite{raajesh2024micap} further unifies identity-aware video caption generation and identity filling within a shared autoregressive framework, enabling both complete caption generation and identity recognition from partially anonymized descriptions. In sports videos, Xi et al.~\cite{xi2025player} introduce player-centric visual features and multimodal prompts to associate basketball-related event descriptions with specific players.

Existing identity-aware video captioning methods are not specifically designed for temporally complete human–object interaction motions involving diverse object categories. In contrast, our task involves a subject interacting with varying objects, requiring the model to jointly reason about identity and HOI motion.

\section{Dataset Processing}

\subsection{Source Datasets}

We use BEHAVE~\cite{bhatnagar2022behave} and InterCap~\cite{huang2024intercap} as the source datasets. Both datasets capture real human-object interactions using synchronized multi-view cameras and provide subject identities and interaction motions. However, they do not provide identity-aware HOI motion captions and therefore cannot be directly used for our task. To facilitate model training for this task, we restructure the source datasets through the following steps.

\subsection{Temporal Video Segmentation}

A temporally complete human-object interaction process serves as the basic data unit in our task. Following the temporal identity-agnostic annotations of InterAct~\cite{xu2025interact}, each synchronized multi-view HOI video is divided according to its annotated start and end timestamps. Every resulting segment is associated with an identity-agnostic motion caption describing the interaction performed within that temporal interval. 

\subsection{Identity-Aware HOI Motion Annotation}

Identity-aware HOI motion captioning requires an explicit correspondence between the physical participant and the performed interaction. Accordingly, the subjects in BEHAVE and InterCap are mapped to a unified set of 18 global identity labels, denoted as \texttt{Sub01} to \texttt{Sub18}. Each label represents one subject and remains consistent across all motion segments and camera views. DeepSeek-V4-Pro~\cite{xu2026deepseek} is then employed to rewrite the identity-agnostic caption of each segment into an identity-aware HOI motion caption. Specifically, the generic subject expression is replaced with the corresponding global identity label, while the principal interaction motion and object are preserved. Finally, all captions are manually inspected and corrected to ensure consistency between the subject identity and interaction semantics.

\subsection{Data Splits}

The data partition is designed to support multi-view feature learning and gallery-based identity-aware HOI motion captioning evaluation. As summarized in Table~\ref{tab:data_stats}, the processed dataset contains 2,073 multi-view HOI segments from 18 subjects. Among them, 1,700 segments from 10 subjects are used for training, while the remaining 373 segments from 8 subjects are assigned to testing. For each test subject, one segment is selected as the gallery, resulting in 8 gallery segments, and the remaining 365 segments form the query set. During training, the model receives multi-view inputs to learn identity and interaction motion features. During testing, each query is processed from a single view, and its identity is determined by matching its feature against the multi-view gallery samples.

Although each query segment contains four synchronized camera views, testing is performed under a single-view setting. Specifically, the four views of each query segment are unfolded and evaluated independently, resulting in $365\times4=1{,}460$ single-view query instances. Captioning and identity-recognition metrics are computed over all 1,460 view-level predictions.

\begin{table}[t]
\centering
\small
\setlength{\tabcolsep}{3pt}
\begin{tabular}{@{}lccc@{}}
\toprule
\textbf{Statistic}
& \textbf{Training}
& \multicolumn{2}{c}{\textbf{Testing}} \\
\cmidrule(lr){3-4}
&
& \textbf{Query}
& \textbf{Gallery} \\
\midrule
Segments      & 1,700 & 365 & 8 \\
Subjects      & 10    & 8   & 8 \\
Views Per Segment & 4     & 4   & 4 \\
\bottomrule
\end{tabular}
\caption{Statistics of the training, query, and gallery sets for identity-aware HOI motion captioning.}
\label{tab:data_stats}
\end{table}

\section{Method}

\begin{figure*}[!t]
\centering
\includegraphics[width=\textwidth]{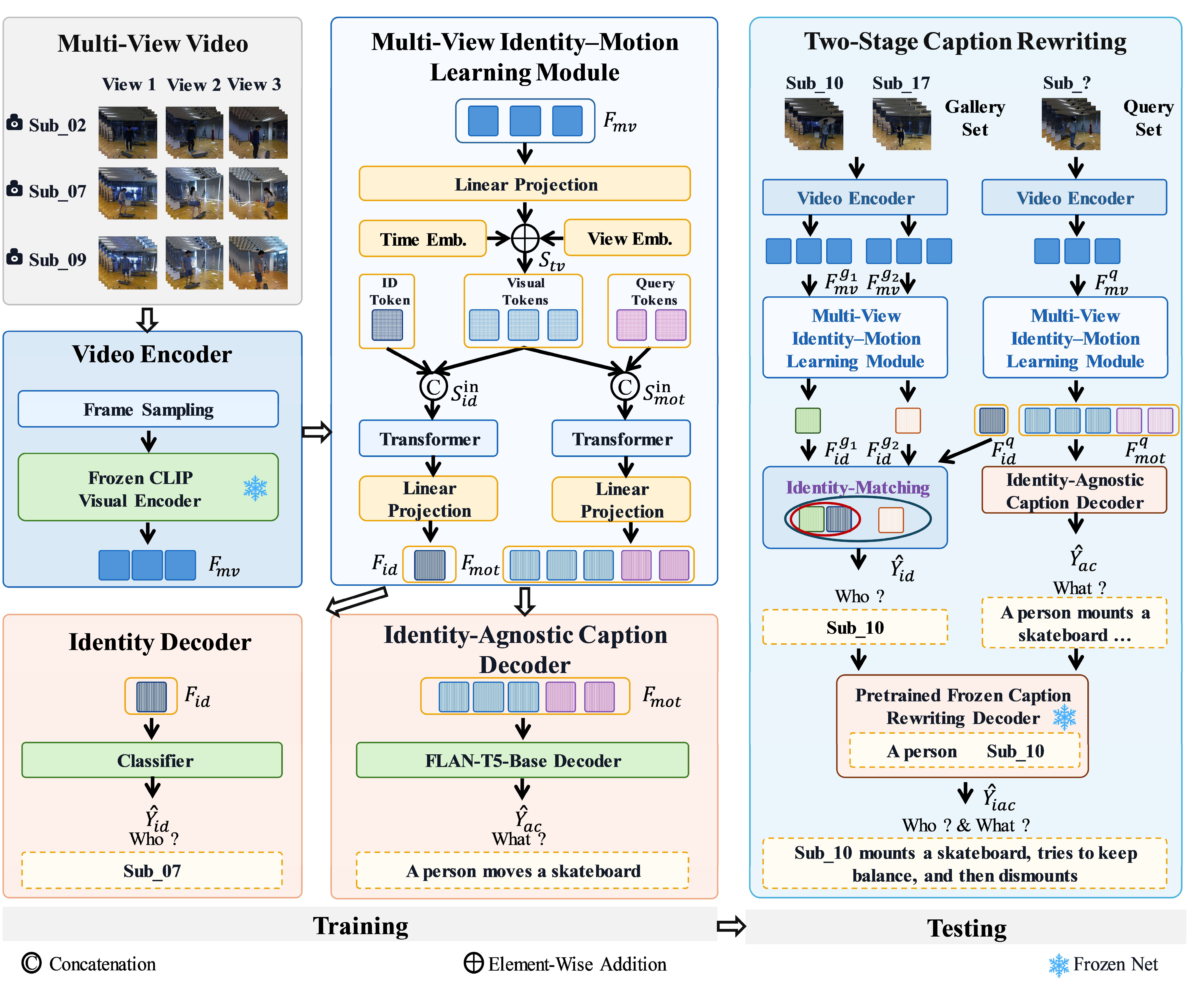}
\caption{Overview of the proposed \textbf{ID-HOINet}. During training, synchronized multi-view videos are processed by a frozen CLIP visual encoder, and the Multi-View Identity-Motion Learning Module learns identity and interaction motion features for identity recognition and identity-agnostic caption generation. During testing, the Two-Stage Caption Rewriting Strategy first determines the query identity through gallery matching and generates an identity-agnostic motion caption, then rewrites the caption with the determined identity using a pretrained frozen Caption Rewriting Decoder.}
\label{fig:method}
\end{figure*}

\subsection{Overview}

During training, ID-HOINet takes synchronized multi-view HOI videos as input. The Video Encoder extracts visual features from sampled frames, which are subsequently processed by the Multi-View Identity-Motion Learning Module (MVIML) to learn identity and interaction motion features. The Identity Decoder determines the subject identity, while the Identity-Agnostic Caption Decoder generates the corresponding caption.

During testing, the proposed Two-Stage Caption Rewriting Strategy (\textbf{TSCR}) operates in two stages. In the first stage, the gallery samples and each single-view query are processed by the shared Video Encoder and Multi-View Identity-Motion Learning Module (MVIML). The query identity is determined through gallery matching, while an identity-agnostic motion caption is generated independently. In the second stage, the determined identity and generated motion caption are integrated by the pretrained frozen Caption Rewriting Decoder to produce the final identity-aware HOI motion caption.
\subsection{Input Representation}
Given a multi-view HOI clip, we first sample frames from all camera views. Each frame corresponds to the same moment observed from different viewpoints. The sampled frames are encoded by a frozen CLIP \cite{radford2021learning} visual encoder to obtain multi-view visual feature $F_{mv}$:
\begin{equation}
F_{mv} =
E_{\mathrm{CLIP}}(\mathcal{I}),
\end{equation}
where $B$, $T$, $D$, and $V$ denote the batch size, temporal length, feature dimension, and number of camera views, respectively. $F_{mv} \in
\mathbb{R}^{B\times T\times V\times D}$ provides a multi-view visual feature for subsequent identity-aware motion captioning learning. 

\subsection{Multi-View Identity-Motion Learning}

Multi-view observations capture consistent subject appearance and human-object motion cues across temporal stages and viewpoints. Based on these physical characteristics, the \textbf{Multi-View Identity-Motion Learning Module (MVIML)} transforms $F_{mv}$ into a shared temporal-view feature and employs two independent branches to learn identity and interaction motion features.

\textbf{Multi-View Representation.}
To distinguish the temporal and viewpoint positions associated with these observations,
MVIML projects $F_{mv}$ into a shared token space and incorporates
temporal $p_{t}^{\mathrm{time}} \in\mathbb{R}^{B\times TV\times D}$ and viewpoint $p_{t}^{\mathrm{view}} \in\mathbb{R}^{B\times TV\times D}$ features:
\begin{equation}
S_{{tv}}
=
\operatorname{Flatten}
\left(
 \operatorname{Linear}(F_{mv})
+
p_{t}^{\mathrm{time}}
+
p_{v}^{\mathrm{view}}
\right).
\end{equation}
where $\operatorname{Linear}_{\mathrm{}}(\cdot)$ denotes a linear projection layer, $S_{{tv}}\in\mathbb{R}^{B\times TV\times D}$ is the temporal-view feature, in which each token represents the interaction observed at a specific temporal stage and camera
viewpoint.

\textbf{Multi-View Identity Learning.}
To learn cross-view identity information, a learnable identity $S_{{id}}^{\mathrm{cls}} \in
\mathbb{R}^{B\times1\times D}$ token is introduced. The identity token represents
global subject characteristics shared across different observations. It is concatenated with the temporal-view
feature to construct the identity-augmented temporal-view token sequence $S_{{id}}^\mathrm{in}\in
\mathbb{R}^{B\times(TV+1)\times D}$:
\begin{equation}
S_{{id}}^\mathrm{in}
=
[S_{{id}}^{\mathrm{cls}};
S_{{tv}}].
\end{equation}
where $[\cdot\,;\,\cdot]$ denotes concatenation along the temporal-view dimension.

Then an identity-specific transformer is introduced.
The transformer models the identity-augmented temporal-view token sequence to capture the identity feature $F_{{id}} \in
\mathbb{R}^{B\times1\times D}$:
\begin{equation}
F_{{id}}
=
\operatorname {Linear}\left(
\operatorname{Transformer}_{{id}}
\left(S_{{id}}^\mathrm{in}\right)
\right)_{\{{cls}\}},
\end{equation}
where $[\cdot]_{\{{cls}\}}$ denotes the selection of the encoded outputs corresponding to the identity token.

\textbf{Multi-View Motion Learning.}
To learn cross-view motion information, a set of learnable motion query tokens $S_{{mot}}^{\mathrm{q}} \in
\mathbb{R}^{B\times N_{{mot}}\times D}$ is also introduced. These tokens are designed to mine key HOI motion features from different
temporal stages and viewpoints. Together with $S_{\mathrm{tv}}$, these
tokens form the motion-augmented temporal-view token sequence $S_{{mot}}^\mathrm{in}\in
\mathbb{R}^{B\times(TV+N_{{mot}})\times D}$:
\begin{equation}
S_{{mot}}^\mathrm{in}
=
[S_{{tv}};
S_{{mot}}^{\mathrm{q}}].
\end{equation}
where $N_{{mot}}$ denotes the number of motion query tokens.

Then, the motion-specific transformer models the relation between the
temporal-view feature and motion query tokens.
The transformer produces the interaction motion feature $F_{\mathrm{mot}}\in
\mathbb{R}^{B\times(TV+N_{{mot}})\times D}$:
\begin{equation}
F_{{mot}}
=
\operatorname {Linear}\left(\operatorname{Transformer}_{{mot}}
\left(S_{{mot}}^\mathrm{in}\right)\right).
\end{equation}

\subsection{Two-Stage Caption Rewriting}

TSCR is applied only during testing. Its first stage uses the trained Identity Decoder and Identity-Agnostic Caption Decoder to retrieve the query identity and generate an identity-agnostic HOI motion caption, respectively. In the second stage, the retrieved identity and generated caption are fed into the Caption Rewriting Decoder to produce the final identity-aware HOI motion caption. The training process of these decoders are detailed in the supplementary material.

\paragraph{Training.}
To obtain the identity and motion outputs required by the first testing stage, we train the Identity Decoder and Identity-Agnostic Caption Decoder in parallel.

The Identity Decoder maps the identity feature to a subject entity $Y_{\mathrm{id}}$ over the identity feature $F_{id}$:
\begin{equation}
Y_{\mathrm{id}}
=
\operatorname{IDD}
\left(
F_{\mathrm{id}}
\right).
\end{equation}

In parallel, the Identity-Agnostic Caption Decoder, initialized from FLAN-T5-Base \cite{chung2024scaling}, maps the interaction motion feature $F_mot$ to an identity-agnostic HOI motion caption:
\begin{equation}
p
\left(
Y_{\mathrm{ac}}
\mid
F_{\mathrm{mot}}
\right)
=
\prod_{l=1}^{L}
p_{\mathrm{IACD}}
\left(
u_l
\mid
u_{<l},
F_{\mathrm{mot}}
\right),
\end{equation}
where $Y_{\mathrm{ac}}=\{u_l\}_{l=1}^{L}$ denotes the ground-truth identity-agnostic HOI motion caption, $u_l$ denotes its $l$-th token, $u_{<l}$ denotes the preceding tokens, and $L$ denotes the caption length.

\paragraph{Two-Stage Testing.}
The test subjects are excluded from the training identity set; therefore, the predicted subject identity is not suitable during testing.

\textbf{(1) First stage.}
Each single-view query video and each reference gallery video are independently processed by the shared identity branch to obtain their identity feature. The query identity feature is compared with each gallery identity feature, and the identity label associated with the most similar gallery sample is retrieved:
\begin{equation}
j^{*}
=
\underset{j\in\{1,\ldots,G\}}
{\operatorname{argmax}}
\;
\operatorname{sim}
\left(
Y_{\mathrm{id}}^{\mathrm{q}},
Y_{\mathrm{id}}^{\mathrm{g}_{j}}
\right),
\end{equation}
\begin{equation}
\hat{Y}_{\mathrm{id}}
=
Y_{\mathrm{id}}^{\mathrm{g}_{j^{*}}},
\end{equation}
where $G$ denotes the number of gallery samples,
$Y_{\mathrm{id}}^{\mathrm{q}}$ denotes the query identity feature,
$Y_{\mathrm{id}}^{\mathrm{g}_{j}}$ denotes the identity feature of the $j$-th gallery sample, and
$\operatorname{sim}(\cdot,\cdot)$ denotes the similarity function.
$Y_{\mathrm{id}}^{\mathrm{g}_{j^{*}}}$ is the identity label of the matched gallery sample, and $\hat{Y}_{\mathrm{id}}$ is the retrieved query identity.

Meanwhile, the Identity-Agnostic Caption Decoder independently generates an HOI motion caption from the query motion feature:
\begin{equation}
\hat{Y}_{\mathrm{ac}}
=
\operatorname{IACD}
\left(
F_{\mathrm{mot}}^{\mathrm{q}}
\right),
\end{equation}
where $F_{\mathrm{mot}}^{\mathrm{q}}$ denotes the interaction motion feature of the query video and $\hat{Y}_{\mathrm{ac}}$ denotes the generated identity-agnostic HOI motion caption. Thus, the first stage obtains the query identity and HOI motion semantics independently.

\textbf{(2) Second stage.}
The pretrained Caption Rewriting Decoder incorporates the retrieved identity into the generated motion caption:
\begin{equation}
\hat{Y}_{\mathrm{iac}}
=
\operatorname{CRD}
\left(
\hat{Y}_{\mathrm{id}},
\hat{Y}_{\mathrm{ac}}
\right),
\end{equation}
where $\hat{Y}_{\mathrm{iac}}$ denotes the final identity-aware HOI motion caption. This stage explicitly associates the retrieved subject identity with the corresponding HOI motion within a unified caption.

\begin{table*}[t]
\centering
\small
\setlength{\tabcolsep}{5pt}
\begin{tabular}{lcccccc}
\toprule
Method

& BLEU-4
& METEOR
& ROUGE-L
& CIDEr 
& ID Acc
& Weighted ID Acc\\
\midrule
%
%
%
%
%
CLIP-Captioner~\cite{yang2022clip}
& 5.28 & 29.28 & 24.21 & 39.33
& {71.57} & 57.50 \\

CARE~\cite{yang2023concept}
& 8.04 & 31.80 & 28.81 & 55.12
& 65.99 & 47.88 \\

NACF~\cite{yang2021non}
& 4.29 & 25.80 & 25.56 & 56.45
& 65.80 & 57.20 \\

CoCap~\cite{shen2023accurate}
& 4.32 & 28.05 & 24.08 & 48.63
& 71.62 & {56.62} \\

SwinBERT~\cite{lin2022swinbert}
& 8.81 & 34.45 & 29.29 & 55.26
& 65.57 & 46.21 \\

\textbf{ID-HOINet}
& \textbf{10.75}
& \textbf{35.99}
& \textbf{33.94}
& \textbf{97.53}
& \textbf{72.94}
& \textbf{58.31} \\
\bottomrule
\end{tabular}

\caption{Quantitative comparison with representative video captioning methods for identity-aware HOI motion captioning. }
\label{tab:test2}
\end{table*}
\section{Experiments}

\subsection{Evaluation Settings}
We compare ID-HOINet with representative video captioning methods to evaluate its effectiveness. Ablation experiments further examine the contributions of each component of ID-HOINet, while qualitative visualizations illustrate the model's ability to associate subject identities with the corresponding HOI motions in generated captions.
\subsection{Metrics}

We evaluate caption quality using four standard metrics: BLEU-4~\cite{papineni2002bleu},
METEOR~\cite{banerjee2005meteor},
ROUGE-L~\cite{lin2004rouge}, and
CIDEr~\cite{vedantam2015cider}. BLEU-4 evaluates n-gram precision up to four-grams, METEOR measures unigram-level semantic alignment, ROUGE-L captures sequence-level similarity based on the longest common subsequence, and CIDEr assesses n-gram caption consensus.

Following NBA-Identity (Xi et al. 2025), we additionally employ ID Accuracy (ID Acc) and Weighted ID Accuracy (Weighted ID Acc) to evaluate identity recognition. ID Acc measures the proportion of query samples whose predicted identity labels match their ground-truth identities by treating all query samples equally. Weighted ID Acc first computes the prediction accuracy for each identity and then averages the identity-level accuracies, thereby reducing the influence of imbalanced query sample distributions across identities.

\subsection{Baselines}

We compare ID-HOINet with five representative video captioning methods: CLIP-Captioner \cite{yang2022clip}, CARE \cite{yang2023concept}, NACF \cite{yang2021non}, CoCap \cite{shen2023accurate}, and SwinBERT \cite{lin2022swinbert}. For a fair comparison, all baseline methods are evaluated under the same evaluation setting.

\begin{table*}[t]
\centering
\small
\setlength{\tabcolsep}{2.5pt}

\begin{tabular}{lcccccc}
\toprule
Variant & BLEU-4 & METEOR & ROUGE-L & CIDEr & ID Acc & Weighted ID Acc  \\
\midrule

ID-HOINet w/o MVIML
& 6.24
& 26.72
& 28.77
& 37.40
& 45.49
& 12.50 \\

ID-HOINet w/o view randomization
& 9.46
& 33.27
& 32.11
& 77.65
& 52.27
& 49.79 \\

ID-HOINet w/ direct identity-aware generation
& \textbf{12.15}
& \textbf{38.19}
& \textbf{34.24}
& 95.41
& 67.39
& 52.75 \\

\textbf{ID-HOINet}
& 10.75
& 35.99
& 33.94
& \textbf{97.53}
& \textbf{72.94}
& \textbf{58.31} \\

\bottomrule
\end{tabular}
\caption{Ablation study of ID-HOINet.}
\label{tab:ablation_test2}
\end{table*}
\subsection{Implementation Details}
We implement ID-HOINet in PyTorch. All video frames are represented by pre-extracted CLIP visual features, and each clip is sampled with 32 temporal steps from 4 camera views. The CLIP feature dimension is 768. In MVIML, we employ 1 learnable identity token and 8 learnable motion query tokens, which are processed by two independent Transformer branches for identity recognition and identity-agnostic captioning, respectively. During training, we apply a view-randomization strategy that randomly keeps 2, 3, or 4 views with probabilities of 0.5, 0.3, and 0.2, respectively, so that the model learns robust multi-view features while remaining compatible with single-view inference. The final identity-aware HOI motion caption is produced by a pretrained FLAN-T5-Base Caption Rewriting Decoder.

The model is trained for 50 epochs using two AdamW optimizers with a weight decay of \(1\times10^{-4}\). The main net use a learning rate of \(5\times10^{-5}\), while the trainable FLAN-T5-Base decoder uses a learning rate of \(1\times10^{-5}\). Both learning rates are linearly warmed up over the first five epochs and then kept constant. The batch size is 32. All experiments are conducted on a single NVIDIA GeForce RTX 4090 GPU.

\subsection{Datasets}
\textbf{BEHAVE}~\cite{bhatnagar2022behave} is a multi-view HOI video dataset captured by four synchronized cameras. It contains 321 video sequences of eight subjects interacting with 20 everyday objects across five indoor environments. Diverse full-body interactions are recorded in RGB video, including both hand-object and foot-object motions.

\textbf{InterCap}~\cite{huang2024intercap} contains 223 multi-view HOI videos captured by six synchronized cameras. It records ten subjects interacting with ten everyday objects of different sizes and affordances. It provides RGB interaction videos depicting diverse full-body interactions, including both hand-object and foot-object motions.

\subsection{Comparison with Captioning Baselines}

Table~\ref{tab:test2} shows that the ID-HOINet consistently outperforms representative video captioning methods across both captioning and identity-matching metrics, demonstrating its effectiveness for identity-aware HOI motion captioning. These improvements indicate that ID-HOINet can generate more accurate interaction captions while maintaining reliable subject identification. Taken together, these results establish ID-HOINet as a strong baseline for future research on this task.

\begin{figure*}[!t]
\centering
\includegraphics[width=\linewidth]{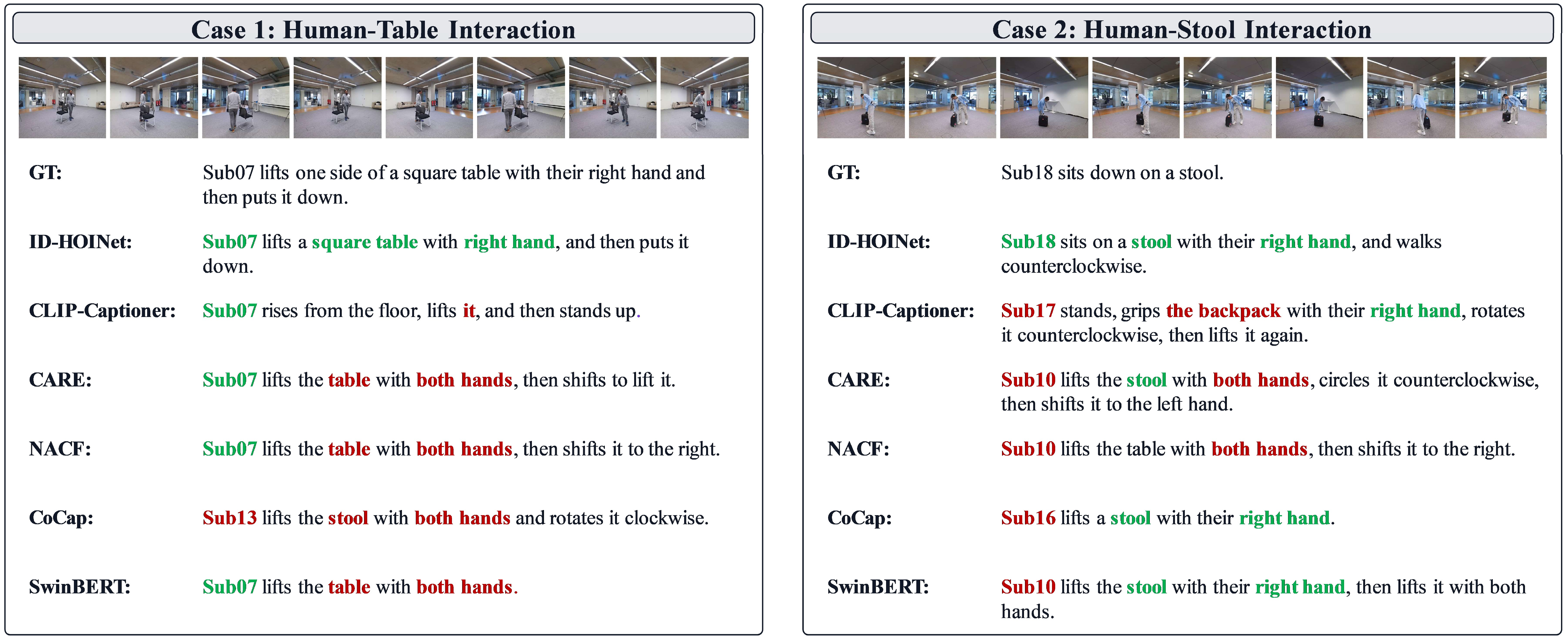}
\caption{Qualitative comparison of identity-aware HOI motion captions generated by ID-HOINet and representative video captioning methods. Each case presents sampled video frames, the ground-truth caption, and generated results, with correct and incorrect identity, object, and motion expressions highlighted in green and red, respectively.}

\label{fig:visual}
\end{figure*}

\subsection{Ablation Study}

Table~\ref{tab:ablation_test2} examines how multi-view identity-motion learning, view-randomized training, and two-stage caption generation contribute to identity matching and HOI motion captioning under single-view testing. Removing MVIML causes substantial degradation across all metrics, particularly ID Acc and Weighted ID Acc, demonstrating the importance of multi-view information for jointly understanding subject identity and HOI motion. 

Removing the view randomization strategy also reduces both captioning and identity-matching performance, indicating that randomized multi-view training improves robustness under single-view testing.

We further evaluate a direct identity-aware HOI motion caption generation strategy. Although this variant achieves higher captioning scores, its ID Acc decreases from 72.94 to 67.39, and its Weighted ID Acc decreases from 58.31 to 52.75. This result reveals a trade-off between textual similarity and identity reliability: a caption may achieve high lexical overlap with the reference while still being associated with an incorrect subject. In identity-aware HOI motion captioning, identity misattribution is particularly detrimental, as assigning an otherwise accurate HOI motion caption to the wrong subject fundamentally compromises the semantic validity of the final prediction. Therefore, we adopt the proposed two-stage generation strategy rather than direct identity-aware caption generation.

%

\subsection{Qualitative Results}

Figure~\ref{fig:visual} presents two representative comparisons of identity-aware HOI motion captions. In Case 1, ID-HOINet correctly identifies Sub07 and accurately describes the interacted object, the right-hand operation, and the complete motion sequence of lifting and putting down the square table. In contrast, the baseline methods either confuse the object, incorrectly describe the use of both hands, omit the subsequent putting-down action, or assign the interaction to the wrong subject. In Case 2, ID-HOINet correctly associates the sitting action with Sub18 and the stool, whereas all comparison methods misidentify the subject and most incorrectly interpret the interaction as lifting an object. Although ID-HOINet introduces additional motion details in this case, it still preserves the correct identity and core HOI semantics.

Overall, these examples show that ID-HOINet achieves better consistency among subject identity, interacted object, and HOI motion than video captioning baselines. The results also indicate that identity-aware HOI motion captioning requires more than lexical similarity: an effective prediction must correctly associate the observed interaction with the corresponding subject.

\section{Limitations}

As an initial study of identity-aware HOI motion captioning, this work has several limitations. First, the experiments are conducted on reorganized BEHAVE and InterCap data containing only 18 subjects and a relatively limited range of objects, actions, and capture environments. The generalization ability of the proposed method to larger subject populations and more diverse real-world scenes therefore remains to be investigated. Second, the current setting focuses on interactions involving a single subject and a single object, without considering more complex multi-person or multi-object interactions. Future work will address these limitations by expanding the data scale and extending the task to more complex interactions.

\section{Conclusion}

This paper introduces Identity-Aware HOI Motion Captioning, which requires a model to generate captions describing both the subject identity and the corresponding human-object interaction motion. We process BEHAVE and InterCap through temporal video segmentation, identity-aware HOI motion caption annotation, and data partitioning. We further propose ID-HOINet, whose Multi-View Identity-Motion Learning Module learns complementary identity and interaction motion features from synchronized multi-view videos. During testing, the Two-Stage Caption Rewriting Strategy first determines the query identity through gallery matching and generates an identity-agnostic motion caption, then rewrites the caption with the determined identity. Experimental results demonstrate the effectiveness of ID-HOINet in associating subject identities with the corresponding HOI motions in generated captions. Future work extends this task to multi-person and multi-object interactions and improves the description of fine-grained motion details.

\bibliography{references}

@inproceedings{tang2021clip4caption,
  title     = {{CLIP4Caption}: {CLIP} for Video Caption},
  author    = {Tang, Mingkang and Wang, Zhanyu and Liu, Zhenhua and Rao, Fengyun and Li, Dian and Li, Xiu},
  booktitle = {Proceedings of the 29th ACM International Conference on Multimedia},
  pages     = {4858--4862},
  year      = {2021},
  doi       = {10.1145/3474085.3479207}
}

@inproceedings{lin2022swinbert,
  title={Swinbert: End-to-end transformers with sparse attention for video captioning},
  author={Lin, Kevin and Li, Linjie and Lin, Chung-Ching and Ahmed, Faisal and Gan, Zhe and Liu, Zicheng and Lu, Yumao and Wang, Lijuan},
  booktitle={Proceedings of the IEEE/CVF conference on computer vision and pattern recognition},
  pages={17949--17958},
  year={2022}
}

@inproceedings{yang2023vid2seq,
  title={Vid2seq: Large-scale pretraining of a visual language model for dense video captioning},
  author={Yang, Antoine and Nagrani, Arsha and Seo, Paul Hongsuck and Miech, Antoine and Pont-Tuset, Jordi and Laptev, Ivan and Sivic, Josef and Schmid, Cordelia},
  booktitle={Proceedings of the IEEE/CVF conference on computer vision and pattern recognition},
  pages={10714--10726},
  year={2023}
}

@inproceedings{wang2024omnivid,
  title={Omnivid: A generative framework for universal video understanding},
  author={Wang, Junke and Chen, Dongdong and Luo, Chong and He, Bo and Yuan, Lu and Wu, Zuxuan and Jiang, Yu-Gang},
  booktitle={Proceedings of the IEEE/CVF conference on computer vision and pattern recognition},
  pages={18209--18220},
  year={2024}
}

@inproceedings{bhatnagar2022behave,
  title={Behave: Dataset and method for tracking human object interactions},
  author={Bhatnagar, Bharat Lal and Xie, Xianghui and Petrov, Ilya A and Sminchisescu, Cristian and Theobalt, Christian and Pons-Moll, Gerard},
  booktitle={Proceedings of the IEEE/CVF conference on computer vision and pattern recognition},
  pages={15935--15946},
  year={2022}
}

@article{huang2024intercap,
  title={InterCap: joint markerless 3D tracking of humans and objects in interaction from multi-view RGB-D images},
  author={Huang, Yinghao and Taheri, Omid and Black, Michael J and Tzionas, Dimitrios},
  journal={International Journal of Computer Vision},
  volume={132},
  number={7},
  pages={2551--2566},
  year={2024},
  publisher={Springer}
}

@inproceedings{ma2018attend,
  title={Attend and interact: Higher-order object interactions for video understanding},
  author={Ma, Chih-Yao and Kadav, Asim and Melvin, Iain and Kira, Zsolt and AlRegib, Ghassan and Graf, Hans Peter},
  booktitle={Proceedings of the IEEE conference on computer vision and pattern recognition},
  pages={6790--6800},
  year={2018}
}

@article{mittal2022savchoi,
  title={SAVCHOI: Detecting suspicious activities using dense video captioning with human object interactions},
  author={Mittal, Ansh and Ghosal, Shuvam and Bansal, Rishibha and Ngyuyen, Dat},
  journal={arXiv preprint arXiv:2207.11838},
  volume={2},
  year={2022}
}

@inproceedings{zhao2025exploring,
  title={Exploring fine-grained human motion video captioning},
  author={Zhao, Bingchan and Liu, Xinyi and Yu, Zhuocheng and Yang, Tongchen and Song, Yifan and Jin, Mingyu and Li, Sujian and Wang, Yizhou},
  booktitle={Proceedings of the 31st International Conference on Computational Linguistics},
  pages={5247--5264},
  year={2025}
}

@inproceedings{song2025towards,
  title={Towards Fine-Grained Human Motion Video Captioning},
  author={Song, Guorui and Wang, Guocun and Huang, Zhe and Lin, Jing and Zhe, Xuefei and Li, Jian and Wang, Haoqian},
  booktitle={Proceedings of the 33rd ACM International Conference on Multimedia},
  pages={846--855},
  year={2025}
}

@inproceedings{xu2025interact,
  title={Interact: Advancing large-scale versatile 3d human-object interaction generation},
  author={Xu, Sirui and Li, Dongting and Zhang, Yucheng and Xu, Xiyan and Long, Qi and Wang, Ziyin and Lu, Yunzhi and Dong, Shuchang and Jiang, Hezi and Gupta, Akshat and others},
  booktitle={Proceedings of the Computer Vision and Pattern Recognition Conference},
  pages={7048--7060},
  year={2025}
}

@article{chung2024scaling,
  title={Scaling instruction-finetuned language models},
  author={Chung, Hyung Won and Hou, Le and Longpre, Shayne and Zoph, Barret and Tay, Yi and Fedus, William and Li, Yunxuan and Wang, Xuezhi and Dehghani, Mostafa and Brahma, Siddhartha and others},
  journal={Journal of Machine Learning Research},
  volume={25},
  number={70},
  pages={1--53},
  year={2024}
}

@inproceedings{papineni2002bleu,
  title={Bleu: a method for automatic evaluation of machine translation},
  author={Papineni, Kishore and Roukos, Salim and Ward, Todd and Zhu, Wei-Jing},
  booktitle={Proceedings of the 40th annual meeting of the Association for Computational Linguistics},
  pages={311--318},
  year={2002}
}

@inproceedings{banerjee2005meteor,
  title={METEOR: An automatic metric for MT evaluation with improved correlation with human judgments},
  author={Banerjee, Satanjeev and Lavie, Alon},
  booktitle={Proceedings of the acl workshop on intrinsic and extrinsic evaluation measures for machine translation and/or summarization},
  pages={65--72},
  year={2005}
}

@inproceedings{lin2004rouge,
  title={Rouge: A package for automatic evaluation of summaries},
  author={Lin, Chin-Yew},
  booktitle={Text summarization branches out},
  pages={74--81},
  year={2004}
}

@inproceedings{vedantam2015cider,
  title={Cider: Consensus-based image description evaluation},
  author={Vedantam, Ramakrishna and Lawrence Zitnick, C and Parikh, Devi},
  booktitle={Proceedings of the IEEE conference on computer vision and pattern recognition},
  pages={4566--4575},
  year={2015}
}

@inproceedings{yang2022clip,
  title={Clip meets video captioning: Concept-aware representation learning does matter},
  author={Yang, Bang and Zhang, Tong and Zou, Yuexian},
  booktitle={Chinese Conference on Pattern Recognition and Computer Vision (PRCV)},
  pages={368--381},
  year={2022},
  organization={Springer}
}

@article{yang2023concept,
  title={Concept-aware video captioning: Describing videos with effective prior information},
  author={Yang, Bang and Cao, Meng and Zou, Yuexian},
  journal={IEEE Transactions on Image Processing},
  volume={32},
  pages={5366--5378},
  year={2023},
  publisher={IEEE}
}

@inproceedings{yang2021non,
  title={Non-autoregressive coarse-to-fine video captioning},
  author={Yang, Bang and Zou, Yuexian and Liu, Fenglin and Zhang, Can},
  booktitle={Proceedings of the AAAI conference on artificial intelligence},
  volume={35},
  number={4},
  pages={3119--3127},
  year={2021}
}

@inproceedings{shen2023accurate,
  title={Accurate and fast compressed video captioning},
  author={Shen, Yaojie and Gu, Xin and Xu, Kai and Fan, Heng and Wen, Longyin and Zhang, Libo},
  booktitle={Proceedings of the IEEE/CVF international conference on computer vision},
  pages={15558--15567},
  year={2023}
}

@inproceedings{radford2021learning,
  title={Learning transferable visual models from natural language supervision},
  author={Radford, Alec and Kim, Jong Wook and Hallacy, Chris and Ramesh, Aditya and Goh, Gabriel and Agarwal, Sandhini and Sastry, Girish and Askell, Amanda and Mishkin, Pamela and Clark, Jack and others},
  booktitle={International conference on machine learning},
  pages={8748--8763},
  year={2021},
  organization={PmLR}
}

@inproceedings{li2024finerehab,
  title={Finerehab: A multi-modality and multi-task dataset for rehabilitation analysis},
  author={Li, Jianwei and Xue, Jun and Cao, Rui and Du, Xiaoxia and Mo, Siyu and Ran, Kehao and Zhang, Zeyan},
  booktitle={Proceedings of the IEEE/CVF conference on computer vision and pattern recognition},
  pages={3184--3193},
  year={2024}
}

@inproceedings{kryeem2023personalized,
  title={Personalized monitoring in home healthcare: An assistive system for post hip replacement rehabilitation},
  author={Kryeem, Alaa and Raz, Shmuel and Eluz, Dana and Itah, Dorit and Hel-Or, Hagit and Shimshoni, Ilan},
  booktitle={Proceedings of the IEEE/CVF International Conference on Computer Vision},
  pages={1868--1877},
  year={2023}
}

@inproceedings{hakim2019mal,
  title={A-mal: Automatic motion assessment learning from properly performed motions in 3d skeleton videos},
  author={Hakim, Tal and Shimshoni, Ilan},
  booktitle={Proceedings of the IEEE/CVF international conference on computer vision workshops},
  pages={0--0},
  year={2019}
}

@inproceedings{doering2022posetrack21,
  title={Posetrack21: A dataset for person search, multi-object tracking and multi-person pose tracking},
  author={Doering, Andreas and Chen, Di and Zhang, Shanshan and Schiele, Bernt and Gall, Juergen},
  booktitle={Proceedings of the IEEE/CVF Conference on Computer Vision and Pattern Recognition},
  pages={20963--20972},
  year={2022}
}

@inproceedings{wang2020combining,
  title={Combining detection and tracking for human pose estimation in videos},
  author={Wang, Manchen and Tighe, Joseph and Modolo, Davide},
  booktitle={Proceedings of the IEEE/CVF Conference on Computer Vision and Pattern Recognition},
  pages={11088--11096},
  year={2020}
}

@inproceedings{liu2025motions,
  title={Motions as queries: One-stage multi-person holistic human motion capture},
  author={Liu, Kenkun and Fu, Yurong and Yuan, Weihao and Lin, Jing and Li, Peihao and Gu, Xiaodong and Qiu, Lingteng and Wang, Haoqian and Dong, Zilong and Han, Xiaoguang},
  booktitle={Proceedings of the Computer Vision and Pattern Recognition Conference},
  pages={17529--17539},
  year={2025}
}

@article{xu2026deepseek,
  title={Deepseek-v4: Towards highly efficient million-token context intelligence},
  author={Xu, Anyi and Lin, Bangcai and Xue, Bing and Wang, Bingxuan and Xu, Bingzheng and Wu, Bochao and Zhang, Bowei and Lin, Chaofan and Dong, Chen and Ling, Chenchen and others},
  journal={arXiv preprint arXiv:2606.19348},
  year={2026}
}

@inproceedings{park2020identity,
  title={Identity-aware multi-sentence video description},
  author={Park, Jae Sung and Darrell, Trevor and Rohrbach, Anna},
  booktitle={European Conference on Computer Vision},
  pages={360--378},
  year={2020},
  organization={Springer}
}

@inproceedings{raajesh2024micap,
  title={Micap: A unified model for identity-aware movie descriptions},
  author={Raajesh, Haran and Desanur, Naveen Reddy and Khan, Zeeshan and Tapaswi, Makarand},
  booktitle={Proceedings of the IEEE/CVF Conference on Computer Vision and Pattern Recognition},
  pages={14011--14021},
  year={2024}
}

@inproceedings{xi2025player,
  title={Player-centric multimodal prompt generation for large language model based identity-aware basketball video captioning},
  author={Xi, Zeyu and Sun, Haoying and Wu, Yaofei and Yan, Junchi and Zhang, Haoran and Wu, Lifang and Wang, Liang and Chen, Changwen},
  booktitle={Proceedings of the IEEE/CVF International Conference on Computer Vision},
  pages={24330--24339},
  year={2025}
}

@inproceedings{zheng2016mars,
  title={Mars: A video benchmark for large-scale person re-identification},
  author={Zheng, Liang and Bie, Zhi and Sun, Yifan and Wang, Jingdong and Su, Chi and Wang, Shengjin and Tian, Qi},
  booktitle={European conference on computer vision},
  pages={868--884},
  year={2016},
  organization={Springer}
}


\end{document}